\RequirePackage{fix-cm}
\documentclass[twocolumn]{svjour3}          
\smartqed  
\usepackage{graphicx}

\usepackage{times}
\usepackage{epsfig}
\usepackage{multirow}

\usepackage{cite}
\usepackage{algorithm, algpseudocode}
\usepackage{caption}
\usepackage{comment}
\usepackage{amsmath}
\usepackage{amssymb}
\usepackage[dvipsnames,svgnames,x11names]{xcolor}
\definecolor{citecolor}{RGB}{119,185,0} 
\usepackage[pagebackref=false,breaklinks=true,letterpaper=true,colorlinks,citecolor=citecolor,bookmarks=false]{hyperref}

\usepackage{pifont}
\usepackage{color}

\usepackage{amsmath}
\DeclareMathOperator*{\argmax}{arg\,max}

\DeclareMathOperator*{\E}{\mathbb{E}}
\def\eg{\emph{e.g.}} 
\def\ie{\emph{i.e.}} 
\def\etal{\emph{et~al.}}

\newlength\savewidth\newcommand\shline{\noalign{\global\savewidth\arrayrulewidth
  \global\arrayrulewidth 1pt}\hline\noalign{\global\arrayrulewidth\savewidth}}
  
\begin{document}

\title{Rectifying Pseudo Label Learning via Uncertainty Estimation \\for 
Domain Adaptive Semantic Segmentation 
}

\author{
  Zhedong Zheng,\quad Yi Yang
}




\institute{
        Zhedong Zheng and Yi Yang are with the Australian Artificial Intelligence Institute (AAII), University of Technology Sydney, NSW, Australia. \email{zhedong.zheng@student.uts.edu.au, yi.yang@uts.edu.au}
}

\date{Received: date / Accepted: date}

\maketitle

\begin{abstract}
This paper focuses on the unsupervised domain adaptation of transferring the knowledge from the source domain to the target domain in the context of semantic segmentation. Existing approaches usually regard the pseudo label as the ground truth to fully exploit the unlabeled target-domain data. 
Yet the pseudo labels of the target-domain data are usually predicted by the model trained on the source domain. Thus, the generated labels 
inevitably contain the incorrect prediction due to the discrepancy between the training domain and the  test domain, which could be transferred to the final adapted model and largely compromises the training process.

To overcome the problem, this paper proposes to explicitly estimate the prediction uncertainty during training to rectify the pseudo label learning for unsupervised semantic segmentation adaptation. Given the input image, the model outputs the semantic segmentation prediction as well as the uncertainty of the prediction. 
Specifically, we model the uncertainty via the prediction variance and involve the uncertainty into the optimization objective.
To verify the effectiveness of the proposed method, we evaluate the proposed method on two prevalent synthetic-to-real semantic segmentation benchmarks, \ie, GTA5 $\rightarrow$ Cityscapes and SYNTHIA $\rightarrow$ Cityscapes, as well as one cross-city benchmark, \ie, Cityscapes $\rightarrow$ Oxford RobotCar. 
We demonstrate through extensive experiments that the proposed approach (1) dynamically sets different confidence thresholds according to the prediction variance, (2) rectifies the learning from noisy pseudo labels, and (3) achieves significant improvements over the conventional pseudo label learning and yields competitive performance on all three benchmarks.

\keywords{ Unsupervised Domain Adaptation \and Domain Adaptive Semantic Segmentation \and Image Segmentation \and Uncertainty Estimation}

\end{abstract}

\section{Introduction}
 Deep neural networks (DNNs)  have been widely adopted in the field of semantic segmentation, yielding the state-of-the-art performance  \cite{liang2017proposal,wei2018revisiting}.
However, recent works show that DNNs are limited in the scalability to the unseen environments, \eg, the testing data collected in rainy days \cite{hendrycks2019benchmarking,wu2019ace}. One straightforward idea is to annotate more training data of the target environment and then re-train the segmentation model. However, semantic segmentation task usually demands dense annotations and it is unaffordable to manually annotate the pixel-wise label for collected data in new environments. 
To address the challenge, the researchers, therefore, resort to unsupervised semantic segmentation adaption, which takes one step closer to real-world practice. In unsupervised semantic segmentation adaptation, two datasets collected in different environments are considered: a labeled source-domain dataset where category labels are provided for every pixel, and an unlabeled target-domain dataset where only provides the collected data without annotations. Compared with the annotated data in the target domain, the unlabeled data is usually easy to collect. Semantic segmentation adaptation aims at leveraging the labeled source-domain data as well as the unlabeled target-domain data to adapt the well-trained model to the target environment.

The main challenge of semantic segmentation adaption is the discrepancy of data distribution between the source domain and the target domain. There are two lines of methods for semantic segmentation adaptation. On one hand, several existing works focus on the domain alignment by minimizing the distribution discrepancy in different levels, such as pixel level~\cite{wu2018dcan,wu2019ace,hoffman2018cycada}, feature level~\cite{huang2018domain,yue2019domain,luo2019significance,zhang2019manifold} and semantic level~\cite{tsai2018learning,tsai2019domain,wang2019class}. Despite great success, this line of work is sub-optimal. Because the alignment objective drives the model to learn the shared knowledge between domains but ignores the domain-specific knowledge. The domain-specific knowledge is one of the keys to the final target, \ie, the model adapted to the target domain. On the other hand, some researchers focus on learning the domain-specific knowledge of the target domain by fully exploiting the unlabeled target-domain data \cite{zou2018unsupervised,zou2019confidence,han2019unsupervised}. Specifically, this line of methods usually adopts the two-stage pipeline, which is similar to the traditional semi-supervised framework~\cite{lee2013pseudo}. The first step is to predict pseudo labels by the knowledge learned from the labeled data, \eg, the model trained on the source domain. The second step is to minimize the cross-entropy loss on the pseudo labels of the unlabeled target-domain data. In the training process, pseudo labels are usually regarded as accurate annotations to optimize the model. 

\begin{figure*}[t]
\begin{center}
     \includegraphics[width=1\linewidth]{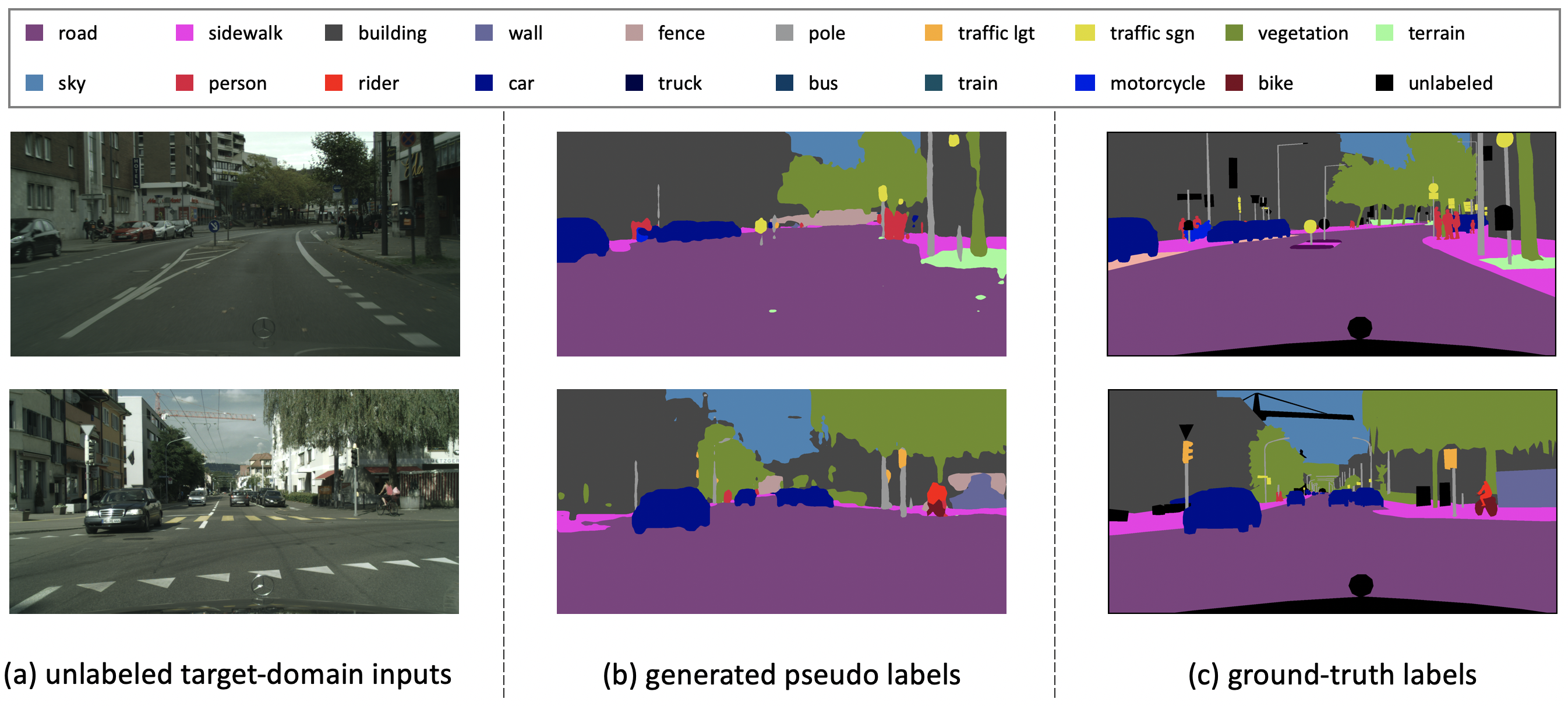}
\end{center} 
      \caption{ Samples of the noisy pseudo labels on Cityscapes~\protect\cite{cordts2016cityscapes}. We leverage the widely-used baseline model~\protect\cite{tsai2018learning} to generate pseudo labels. Despite the large area of correct prediction, the pseudo labels still suffer from the data distribution biases, and inevitably contains incorrect predictions. (Best viewed in \emph{color})}
      \label{fig:pseudo}
\end{figure*}

However, one inherent problem exists in the pseudo label based scene adaptation approaches. Pseudo labels usually suffer from the noise caused by the model trained on different data distribution (see Figure~\ref{fig:pseudo}). The noisy label could compromise the subsequent learning. Although some existing works \cite{zou2018unsupervised,zou2019confidence} have proposed to manually set the threshold to neglect the low-confidence pseudo labels, it is still challenging in several aspects: 
First, the value of the threshold is hard to be determined for different target domain. It depends on the similarity of the source domain and target domain, which is hard to estimate in advance. 
Second, the value of the threshold is also hard to be determined for different categories. For example, the objectives, such as traffic signs, have rarely appeared in the source domain. The overall confidence score for the rare category is relatively low. The high threshold may ignore the information of rare categories.
Third, the threshold is also related to the location of the pixel. For example, the pixel in the center of objectives, such as cars, is relatively easy to predict, while the pixel on the objective edge usually faces ambiguous predictions. It reflects that the threshold should not only consider the confidence score but also the location of the pixel. In summary, every pixel in the segmentation map needs to be treated differently. The fixed threshold is hard to match the demand.

To address the mentioned challenges, we propose one simple and effective method for semantic segmentation adaption via modeling uncertainty, which could provide the pixel-wise threshold for the input image automatically.
Without introducing extra parameters or modules,  we formulate the uncertainty as the prediction variance. The prediction variance reflects the model uncertainty towards the prediction in a bootstrapping manner. 
Meanwhile, we explicitly involve the variance into the optimization objective, called variance regularization, which works as an automatic threshold and is compatible with the standard cross-entropy loss. 
The automatic threshold rectifies the learning from noisy labels and ensures the training in a coherent manner. Therefore, the proposed method could effectively exploit the domain-specific information offered by pseudo labels and takes advantage of the unlabeled target-domain data.

In a nutshell, our contributions are as follows:
\begin{itemize}
\item To our knowledge, we are among the first attempts to exploit the uncertainty estimation and enable the automatic threshold to learn from noisy pseudo labels. This is in contrast to most existing domain adaptation methods that directly utilize noisy pseudo labels or manually set the confidence threshold. 
\item Without introducing extra parameters or modules, we formulate the uncertainty as the prediction variance. Specifically, we introduce a new regularization term, variance regularization, which is compatible with the standard cross-entropy loss. The variance regularization works as the automatic threshold, and rectifies the learning from noisy pseudo labels.
\item We verify the proposed method on two synthetic-to-real benchmarks and one cross-city benchmark. The proposed method has achieved significant improvements over the conventional pseudo label learning, yielding competitive performance to existing methods.
 \end{itemize}

\section{Related work}
\subsection{Semantic Segmentation Adaptation}
The main challenge in unsupervised domain adaptation is different data distribution between the source domain and the target domain \cite{fu2015transductive,wang2018transferable,li2020csrl,li2020metaparsing,kang2020adversarial}. To deal with the challenge, some pioneering works \cite{hoffman2018cycada,wu2018dcan} propose to transfer the visual style of the source-domain data to the target domain. In this way, the model could be trained on the labeled data with the target style. Similarly, some recent works leverage Adversarial Domain Adaptation~\cite{tzeng2015simultaneous,ganin2015unsupervised,luo2020adversarial} to transfer the source-domain images or features to multiple domains and intend to learn the domain-invariant features \cite{wu2019ace,yue2019domain}. Furthermore, some works focus on the alignment among the middle activation of neural networks. Luo \etal \cite{luo2019significance,luo2019taking} utilize the attention mechanism to refine the feature alignment. Instead of modifying the visual appearance, the alignment between the high-level semantic features also attracts a lot of attention. Tsai \etal \cite{tsai2018learning,tsai2019domain} propose to utilize the discriminator to demand the similar semantic outputs between two domains.
In summary, this line of methods focuses on the alignment, learning the shared knowledge between the source and target domains. However, the domain-specific information is usually ignored, which is one of the keys to the adaptation in the target environment. Therefore, in this paper, we resort to another line of methods, which is based on pseudo label learning.

\subsection{Pseudo label learning}
Another line of semantic segmentation adaptation approaches utilizes the pseudo label to adapt the model to target domain \cite{zou2018unsupervised,zou2019confidence,zheng2019unsupervised}. The main idea is close to the conventional semi-supervised learning approach, entropy minimization, which is first proposed to leverage the unlabeled data \cite{grandvalet2005semi}. Entropy minimization encourages the model to give the prediction with a higher confidence score.
In practice, Reed \etal \cite{reed2014training} propose bootstrapping via entropy minimization and show the effectiveness on the object detection and emotion recognition. Furthermore, Lee \etal \cite{lee2013pseudo}
exploit the trained model to predict pseudo labels for the unlabeled data, and then fine-tune the model as supervised learning methods to fully leverage the unlabeled data. Recently, Pan \etal \cite{pan2019transferrable} utilize the pseudo label learning to minimize the distribution of target-domain data with the source-domain prototypes. For unsupervised semantic segmentation, Zou \etal \cite{zou2019confidence,zou2018unsupervised} introduce the pseudo label strategy to the semantic segmentation adaptation and provide one comprehensive analysis on the regularization terms. In a similar spirit, Zheng \etal \cite{zheng2019unsupervised} also apply the pseudo label to learn the domain-specific features, yielding competitive results. However, one inherent weakness of the pseudo label learning is that the pseudo label usually contains noisy predictions. Despite the fact that most pseudo labels are correct, wrong labels also exist, which could compromise the subsequent training. If the model is fine-tuned on the noisy label, the error would also be transferred to the adapted model. 
Different from existing works, we do not treat the pseudo labels equally and intend to rectify the learning from noisy labels. The proposed method explicitly predict the uncertainty of pseudo labels, when fine-tuning the model. The uncertainty could be regarded as an automatic threshold to adjust the learning from noisy labels.

\subsection{Co-training}
Co-training is a semi-supervised learning method, which demands two classifiers to learn complementary information \cite{blum1998combining}. Some domain adaptation works also explore a similar learning strategy. \cite{saito2018maximum,luo2019taking} explicitly maximizes the discrepancy of two classifiers by introducing one extra loss, i.e., the $L_{adv}$ in \cite{saito2018maximum} and the $L_{weight}$ in \cite{luo2019taking}, to obtain complementary classifiers. \cite{saito2018maximum} minimizes the feature discrepancy via adversarial training. Similarly, \cite{luo2019taking} apply the classifier discrepancy on the discriminator loss to stabilize the training.
In contrast, the proposed method enables the classifier discrepancy in nature, since we deploy two classifiers on different intermediate layers. We do not introduce such loss to encourage the classifier discrepancy. Otherwise, every pseudo label will be high-uncertainty. For instance, if the two classifiers output one identical category prediction, we will not punish the network. In contrast, \cite{saito2018maximum} will punish the classifiers for enabling adversarial training. Besides, \cite{saito2018maximum,luo2019taking} still use conventional segmentation loss and do not deal with noisy labels, when the proposed method uses the classifier discrepancy to rectify the pseudo label learning on segmentation.

\subsection{Uncertainty Estimation}
To address the noise, existing works have explored the uncertainty estimation from different aspects, such as the input data, the annotation and the model weights. In this work, we focus on the annotation uncertainty. Our target is to learn a model that could predict whether the annotation is correct, and learn from noisy pseudo labels. Among existing works, Bayesian networks are widely used to predict the uncertainty of weights in the network \cite{nielsen2009bayesian}. 
In a similar spirit, Kendall \etal \cite{kendall2017uncertainties} apply the Bayesian theory to the prediction of computer vision tasks, and intend to provide not only the prediction results but also the confidence of the prediction. 
Further, Yu \etal \cite{yu2019robust} explicitly model the uncertainty via an extra auxiliary branch, and involve the random noise into training. The model could explicitly estimate the feature mean as well as the prediction variance. 
Inspired by the above-mentioned works, we propose to leverage the prediction variance to formulate the uncertainty. There are two fundamental differences between previous works and ours: (1) We do not introduce extra modules or parameters to simulate the noise. Instead, we leverage the prediction discrepancy within the segmentation model. 
(2) We explicitly involve the uncertainty into the training target and adopt the adaptive method to learn the pixel-wise uncertainty map automatically. The proposed method does not need manually setting the threshold to enforce the pseudo label learning.

\section{Methodology}
In Section \ref{sec:problem}, we first provide the problem definition and denotations. We then revisit the conventional domain adaption method based on the pseudo label and discuss the limitation of the pseudo label learning (see Section \ref{sec:revisit}). To deal with the mentioned limitations, we propose to leverage the uncertainty estimation. In particular, we formulate the uncertainty as the prediction variance and provide one brief definition in Section \ref{sec:ue}, followed by the proposed variance regularization, which is compatible with the standard cross-entropy loss in Section \ref{sec:vr}. Besides, the implementation details are provided in Section \ref{sec:detail}.

\subsection{Problem Definition} \label{sec:problem}
Given the labeled dataset $X_s = \{x^i_s\}^M_{i=1}$ from the source domain and the unlabeled dataset $X_t = \{x_t^j\}^N_{j=1}$ from the target domain, semantic segmentation adaptation intends to learn the projection function $F$, which maps the input image $X$ to the semantic segmentation $Y$. $M$ and $N$ denote the number of the labeled data and the unlabeled data. The source-domain semantic segmentation label $Y_s=\{y^i_s\}^M_{i=1}$ is provided for every labeled data of the source domain $X_s$, while the target-domain label $Y_t=\{y^j_t\}^N_{j=1}$ remains unknown during the training. The aim of unsupervised domain adaptation is to estimate the model parameter $\theta_t$, which could minimize the prediction bias on the target-domain inputs:
\begin{align}
Bias(p_t) &= \E [F(x_t^j|\theta_t) - p_t^j],
\label{eq:bias}
\end{align}
where $p_t$ is the ground-truth class probability of target data. Ideally, $p_t^j$ is one-hot vector and the maximum value of $p_t^j$ is $1$. The ground-truth label $y^j_t = \argmax p^j_t $. 
In contrast, $F(x_t^j|\theta_t)$ is the predicted probability distribution of $x_t^j$. When we minimize the prediction bias in Equation~\ref{eq:bias}, the discrepancy between predicted results and the ground-truth probability is minimized. 

\subsection{Pseudo Label Learning Revisit} \label{sec:revisit}

Pseudo label learning is to leverage the pseudo label to learn from the unlabeled data. The common practice contains two stages. The first stage is to generate the pseudo label for the unlabeled target-domain training data. 
The pseudo labels could be obtained via the model trained on source-domain data: $\hat{y}^j_t = \argmax  F(x^j_t|\theta_s)$.
We note that $\theta_s$ is the model parameters learned from the source-domain training data. Therefore, the pseudo labels $\hat{y}_t$, are not accurate in nature due to different data distribution between $X_s$ and $X_t$. We denote $\hat{p}^j_t $ as the one-hot vector of $\hat{y}^j_t $. If the class index $c$ equals to $\hat{y}^j_t$, $\hat{p}^j_t(c) = 1$ else $\hat{p}^j_t(c) = 0$. 
The second stage of pseudo learning is to minimize the prediction bias. We could formulate the bias as the  similar style of Equation~\ref{eq:bias}:
\begin{align}
Bias(p_t) &= \E [F(x_t^j|\theta_t) - \hat{p}_t^j] + \E [\hat{p}_t^j - p_t^j].
\label{eq:ps}
\end{align}
The first term is the difference between the prediction and the pseudo label, while the second term is the error between the pseudo label and the ground-truth label. When fine-tuning the model in the second stage, we fix the pseudo label. Therefore, the second term is one constant. Existing methods usually optimize the first term as the pretext task. It equals to considering the pseudo labels $\hat{p}_t$ as true labels. Existing methods train the model parameter $\theta_t$ to minimize the bias between the prediction and pseudo labels. In practice, the cross-entropy loss is usually adopted \cite{zou2018unsupervised,zou2019confidence,zheng2019unsupervised}. The objective could be formulated as:
\begin{equation}
    L_{ce} = \E [-\hat{p}_t^j \log F(x^j_t|\theta_t) ].
\end{equation}

\noindent\textbf{Discussion.}  There are two advantages of pseudo label learning : First, the model is only trained on the target-domain data. The training data distribution is close to the testing data distribution, minoring the input distribution discrepancy. Second, despite the domain discrepancy,  most pseudo labels are correct. Theoretically, the fine-tuned model could arrive the competitive performance with the fully-supervised model.
However, one inherent problem exists that the pseudo label inevitably contains noise. The wrong annotations are transferred from the source model to the final model. Noisy pseudo label could largely compromise the training.



\begin{figure}[t]
\begin{center}
     \includegraphics[width=1\linewidth]{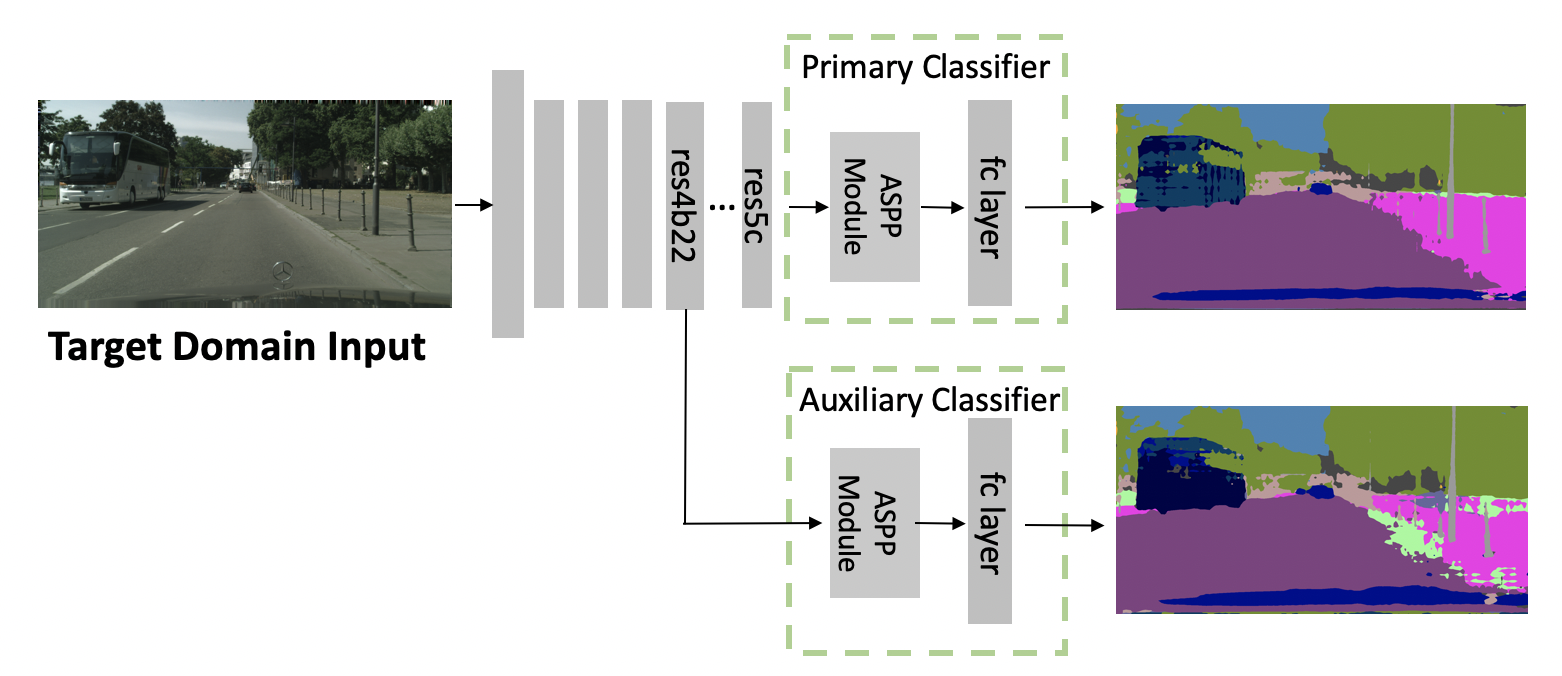}
\end{center} 
\vspace{-.1in}
      \caption{ Illustration of the two-classifier model based on Deeplab-v2 \cite{chen2017deeplab}, which adopts ResNet-101 \cite{he2016deep} as backbone. We follow the previous works \cite{zhao2017pyramid,tsai2018learning,tsai2019domain,luo2019significance,luo2019taking,zheng2019unsupervised} to add an auxiliary classifier with the similar structure as the primary classifier. The auxiliary classifier takes the activation of the shallow layer \emph{res4b22} as the input, while the primary classifier leverages that of \emph{res5c}.
      The ASPP module denotes Atrous Spatial Pyramid Pooling layer \cite{chen2017deeplab}, and the fc layer denotes the fully-connected layer. The original goal of two-classifier model is to evade the problem of gradient vanishing and help the training. In this work, we take one step further to leverage the prediction discrepancy of two classifiers as the uncertainty estimation.
      }
      \label{fig:structure}
\end{figure}

\subsection{Uncertainty Estimation} \label{sec:ue}

To address the label noise, we model the uncertainty of the pseudo label via the prediction variance. 
Intuitively, we could formulate the variance of the prediction as: 
\begin{align}
Var(p_t) &= \E [(F(x_t^j|\theta_t) - p_t^j)^2].
\end{align}
Since $p_t$ remains unknown, one naive way is to utilize the pseudo label $\hat{p}_t$ to replace the $p_t$. The variance could be approximated as:
\begin{align}
Var(p_t)  \approx \E [(F(x_t^j|\theta_t) - \hat{p}_t^j)^2].
\label{eq:fake}
\end{align}
However, in Equation~\ref{eq:ps}, we have pushed $F(x_t^j|\theta_t)$ to $\hat{p}_t$. When optimizing the prediction bias, the variance in Equation~\ref{eq:fake} will also be minimized. It could not reflect the real prediction variance during training. In this paper, therefore, we adopt another approximation as: 
\begin{align}
Var(p_t)  \approx \E [(F(x_t^j|\theta_t) - F_{aux}(x_t^j|\theta_t))^2],
\end{align}
where $F_{aux}(x_t|\theta_t)$ denotes the auxiliary classifier output of the segmentation model. As shown in Figure \ref{fig:structure}, we adopt the widely-used two-classifier model, which contains one primary classifier as well as one auxiliary classifier. We note that the extra auxiliary classifier could be viewed as a free lunch since most segmentation models, including PSPNet \cite{zhao2017pyramid} and the modified DeepLab-v2 in \cite{tsai2018learning,tsai2019domain,luo2019significance,zheng2019unsupervised},  contain the auxiliary classifier to solve the gradient vanish problem \cite{he2016identity} and help the training. In this paper, we further leverage the auxiliary classifier to estimate the variance. In practice, we utilize the KL-divergence of two classifier predictions as the variance:
\begin{align}
D_{kl} = \E [F(x_t^j|\theta_t) \log (\frac{F(x_t^j|\theta_t)}{F_{aux}(x_t^j|\theta_t)})],
\label{eq:var_final}
\end{align}
If two classifiers provide two different class predictions, the approximated variance will obtain one large value. It reflects the uncertainty of the model on the prediction. Besides, it is worthy to note that the proposed variance in Equation~\ref{eq:var_final} is independent with the pseudo label $\hat{p_t}$.

\noindent\textbf{Discussion:} \emph{What leads to the discrepancy of the primary classifier and the auxiliary classifier?} First of all, the main reason is different receptive fields. As shown in Figure \ref{fig:structure}, the auxiliary classifier is located at the relatively shallow layer, when the primary classifier learns from the deeper layer. The input activation is different between two classifiers, leading to the prediction difference. Second, the two classifiers have not been trained on the target-domain data. Therefore, both classifiers may have different biases to the target-domain data. Third, we apply the dropout function \cite{srivastava2014dropout} to two classifiers, which also could lead to the different prediction during training. The prediction discrepancy helps us to estimate the uncertainty.

\begin{algorithm}[t]
\small
\caption{Training Procedure of the Proposed Method}
\label{alg:RECT}
\begin{algorithmic}[1]
\Require The target domain dataset $X_t=\{x_t^j\}^N_{j=1}$; The generated pseudo label $\hat{Y}_t=\{\hat{y}_t^j\}^N_{j=1}$; 
\Require The source-domain parameter $\theta_s$; The iteration number $T$.
\State Initialize $\theta_t = \theta_s$;
\For {$iteration = 1$ to $T$}
\State Input $x_t^j$ to $F(\cdot|\theta_t)$, extract the prediction of two classifiers, calculate the prediction variance according to Equation \ref{eq:var_final}:
\vspace{-1.5ex}
\begin{equation}
D_{kl} = \E [F(x_t^j|\theta_t) \log (\frac{F(x_t^j|\theta_t)}{F_{aux}(x_t^j|\theta_t)})].
\end{equation}
\vspace{-2ex}
\State We fix the prediction variance, and calculate the original cross-entropy loss according to Equation \ref{eq:ps}, where $\hat{p}_t^j$  is the one-hot vector of the pseudo label $\hat{y}_t^j$: 
\vspace{-1.5ex}
\begin{equation}
L_{ce} = \E [-\hat{p}_t^j \log F(x^j_t|\theta_t) ].
\end{equation}
\State We combine the prediction variance with the conventional objective to obtain the rectified objective. Update the $\theta_t$ according to Equation \ref{eq:rectified}:
\vspace{-1.5ex}
\begin{equation}
L_{rect} =  \E [  exp\{-D_{kl}\} L_{ce} +  D_{kl} ]
\end{equation}
\vspace{-4ex}
\EndFor \\
\Return $ \theta_t $.
\end{algorithmic}
\end{algorithm}

\subsection{Variance Regularization}\label{sec:vr}
In this paper, we propose the variance regularization term to rectify the learning from noisy labels. It leverages the approximated variance introduced in Section \ref{sec:ue}. 
The rectified objective could be formulated as:
\begin{align}
L_{rect} =  \E [ \frac{1}{Var(p_t)} Bias(p_t) +  Var(p_t) ]
\end{align} 
It is worthy to note that we do not intend to minimize the prediction bias under all conditions. If the prediction variance has received one large value, we will not punish the prediction bias $Bias(p_t)$. Meanwhile, to prevent that the model predicts the large variance all the time, as a trade-off, we introduce the regularization term via adding $Var(p_t)$. 
Besides, since $Var(p_t)$ could be zero, it may lead to the problem of dividing by zero. To stabilize the training, we adopt the policy in \cite{kendall2017uncertainties} that replace $1/Var$ as $exp(-Var)$. Therefore, the loss term could be rewritten with the approximated terms as:
\begin{align}
L_{rect} =  \E [  exp\{-D_{kl}\} L_{ce} +  D_{kl} ].
\label{eq:rectified}
\end{align} 
The training procedure of the proposed method is summarized in Algorithm \ref{alg:RECT}. In practice, we utilize the parameter $\theta_s$ learned in the source-domain dataset to initialize the $\theta_t$. In every iteration, we calculate the prediction variance as well as the cross-entropy loss for the given inputs. We utilize the $L_{rect}$ to update the $\theta_t$. The training cost of the rectified objective approximately equals to the conventional pseudo label learning, since no extra modules are introduced.

\begin{figure}[t]
\begin{center}
     \includegraphics[width=1\linewidth]{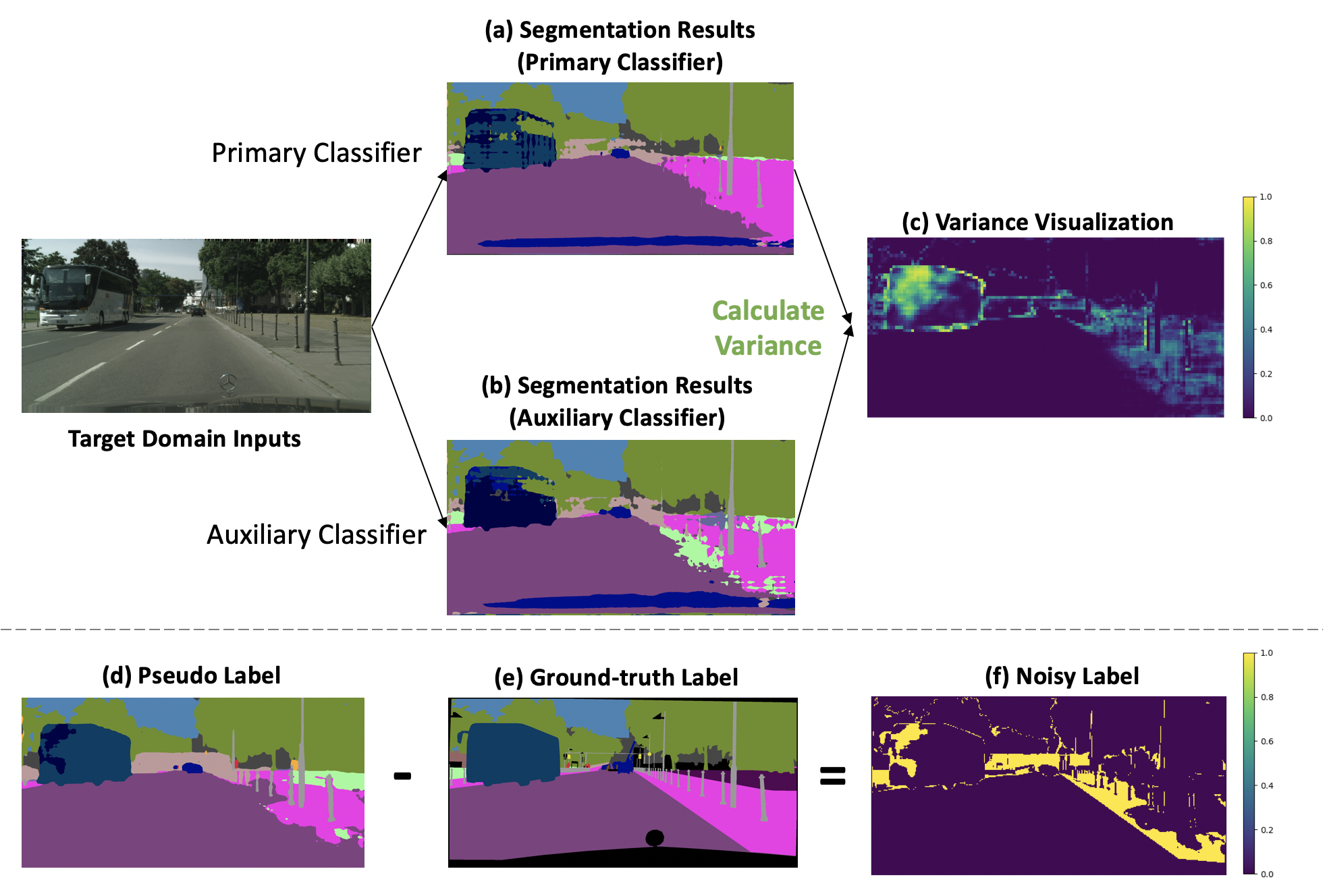}
\end{center} 
      \caption{ Illustration of the prediction variance between two classifiers, \ie, the primary classifier and the auxiliary classifier. The areas, where have ambiguous predictions, obtain large value of the prediction variance. Meanwhile, we could observe that the high-variance area has considerable overlaps with the noise in the pseudo label. (Best viewed in \emph{color}) }
      \label{fig:variance}
\end{figure}

\noindent\textbf{Discussion:} \emph{What are the advantages of the proposed variance regularization?} 
First, the proposed variance regularization does not introduce extra parameters or modules to model the uncertainty. Different from \cite{yu2019robust}, we do not explicitly introduce the Gaussian noise or extra branches. Instead, we leverage the prediction variance of the model itself. Second, the proposed variance regularization has good scalability.
If the variance equals to zero, the optimization loss degrades to the objective of the conventional pseudo learning and the model will focus on minimizing the prediction bias only. In contrast, when the value of variance is high, the model is prone to neglect the bias and skip ambiguous pseudo labels; 
Third, the proposed variance regularization has the same shape of the prediction, and could works as the pixel-wise threshold of the pseudo label. As shown in Figure~\ref{fig:variance}, we could observe that the noise usually exists in the area with high variance. The proposed rectified loss assigns different thresholds to different areas. For example, for the location with coherent predictions, the variance regularization drives the model trust pseudo labels. For the area with ambiguous predictions, the variance regularization drives the model to neglect pseudo labels. Different from existing works that set the unified threshold for all training samples, the proposed pseudo label could provide more accurate and adaptive threshold for every pixel.  

\subsection{Implementation} \label{sec:detail}
\noindent\textbf{Network Architecture.} In this work, we utilize the widely-used Deeplab-v2 \cite{chen2017deeplab} as the baseline model, which adopts the ResNet-101 \cite{he2016deep} as the backbone model. We follow most existing works \cite{tsai2018learning,tsai2019domain,luo2019significance,luo2019taking,zheng2019unsupervised} to add one auxiliary classifier. The auxiliary classifier has similar structure with the primary classifier, including one Atrous Spatial Pyramid Pooling (ASPP) module \cite{chen2017deeplab} and one fully-connected layer. The auxiliary classifier is added after the \emph{res4b22} layer. We also insert the dropout layer \cite{srivastava2014dropout} before the fully-connected layer, and the dropout rate is $0.1$. 

\noindent\textbf{Pseudo Label.} To verify the effectiveness of the proposed method, we deploy two existing methods, \ie, AdaptSegNet \cite{tsai2018learning} and MRNet \cite{zheng2019unsupervised}, to generate the pseudo labels of the target-domain dataset.
\begin{itemize}
\item AdaptSegNet \cite{tsai2018learning} is one widely-adopted baseline model, which utilize the adversarial training to align the semantic outputs. 
\item MRNet \cite{zheng2019unsupervised} is one recent work, which leverages the memory module to regularize the model training, especially for the target-domain data. 
\end{itemize}
Specifically, MRNet arrives superior performance to AdaptSegNet in terms of mIoU on three benchmarks. Therefore, if not specific, we adopt the pseudo label generated by the stronger baseline, \ie, MRNet. \textbf{It is worth mentioning that we do not use source-domain training data. In practice, we fine-tune the model only on the target-domain training data with pseudo labels.}

\noindent\textbf{Training Details.} 
The input image is resized to $1280\times640$ with scale jittering from $[0.8,1.2]$, and then we randomly crop $512 \times 256$ for training. Horizontal flipping is applied with the possibility of $50\%$. We train the model with mini-batch size of $9$, and the parameters of batch normalization layers are also fine-tuned. The learning rate is set to $0.0001$.
Following \cite{zhao2017pyramid,zhang2019dual,zhang2020dynamic}, we deploy the ploy learning rate policy by multiplying the factor $(1-\frac{iter}{total-iter})^{0.9}$. The total iteration is set as $100k$ iterations and we adopt the early-stop strategy. We stop the training after 50k iterations. When inference, we follow \cite{zheng2019unsupervised} to combine the output of both classifier as the final result. $Output = \argmax(F(x_t^j|\theta_t) + 0.5F_{aux}(x_t^j|\theta_t))$.
Our implementation is based on Pytorch \cite{paszke2017automatic}.

\setlength{\tabcolsep}{5pt}
\begin{table}
\footnotesize
\begin{center}
\begin{tabular}{l|c|c|c|c}
\shline
Datasets & GTA5 & SYNTHIA & Cityscapes & Oxford RobotCar\\
\hline
\#Train & 24,966 & 9,400 & 2,975 & 894\\
\#Test  & - & - & 500 & 271 \\
\#Category & 19 & 16 & 19 & 9 \\
Synthetic & \checkmark & \checkmark & $\times$ & $\times$ \\
\shline
\end{tabular}
\end{center}
\caption{List of categories and number of images in four datasets, \ie, GTA5~\cite{richter2016playing}, SYNTHIA~\cite{ros2016synthia}, Cityscapes~\cite{cordts2016cityscapes} and Oxford RobotCar~\cite{RobotCarDatasetIJRR}.
}
\label{table:Dataset}
\end{table}

\section{Experiment} \label{experiments}
\subsection{Datasets and Evaluation Metric}
\noindent\textbf{Datasets.} To simplify, we denote the test setting as A $\rightarrow$ B, where A represents the labeled source domain and B denotes the unlabeled target domain. We evaluate the proposed method on two widely-used synthetic-to-real benchmarks: \ie, GTA5~\cite{richter2016playing}$\rightarrow$Cityscapes~\cite{cordts2016cityscapes} and SYNTHIA5~\cite{ros2016synthia}$\rightarrow$Cityscapes~\cite{cordts2016cityscapes}. Both source dataset, \ie, GTA5 and SYNTHIA are the synthetic datasets, and the corresponding annotation is easy to obtain. 
Specifically, the GTA5 dataset is collected from a video game, which contains $24,966$ images for training. The SYNTHIA dataset is rendered from a virtual city and comes with pixel-level segmentation annotations, containing $9,400$ training images. 
The realistic dataset, Cityscapes, collect street-view scenes from $50$ different cities, which contains $2,975$ training images and $500$ images for validation. 
Besides, we also evaluate the performance on the cross-city benchmark, \ie, Cityscapes~\cite{cordts2016cityscapes}$\rightarrow$Oxford RobotCar~\cite{RobotCarDatasetIJRR}. 
We utilize the annotation of Cityscapes training images in this setting. The Oxford RobotCar dataset serves as the unlabeled target domain, containing $894$ training images and $271$ validation images. We note that this setting is challenging in different weather conditions. Oxford RobotCar is collected in the rainy days, while the Cityscapes dataset is mostly collected in the sunny days. 
The differences between datasets are listed in Table \ref{table:Dataset}.

\noindent\textbf{Evaluation Metric.} We report pre-class IoU and mean IoU over all classes. For SYNTHIA $\rightarrow$ Cityscapes, due the limited annotated classes in the source dataset, we report the results based on 13 categories as well as 16 categories with three small-scale categories. For Cityscapes $\rightarrow$ Oxford RobotCar, we follow the setting in \cite{tsai2019domain} and report 9 pre-class IoU as well as the mIoU accuracy.

\begin{table*}[!t]
	\centering
	\resizebox{\linewidth}{!}{
	\begin{tabular}{c|ccccccccccccccccccc|c}
		\shline
		Method & Road & SW & Build & Wall & Fence & Pole & TL & TS & Veg. & Terrain & Sky & PR & Rider & Car & Truck & Bus & Train & Motor & Bike & mIoU\\
		\shline
		\hline
		Source & 75.8 & 16.8 & 77.2 & 12.5 & 21.0 & 25.5 & 30.1 & 20.1 & 81.3 & 24.6 & 70.3 & 53.8 & 26.4 & 49.9 & 17.2 & 25.9 & 6.5 & 25.3 & 36.0 & 36.6\\
		AdaptSegNet~\cite{tsai2018learning} & 86.5 & 36.0 & 79.9& 23.4 & 23.3 & 23.9 & 35.2 & 14.8 & 83.4 & 33.3 & 75.6 & 58.5 & 27.6 & 73.7 & 32.5 & 35.4 & 3.9 & 30.1 & 28.1 & 42.4\\ 
		SIBAN \cite{luo2019significance} & 88.5 & 35.4 & 79.5 & 26.3 & 24.3 & 28.5 & 32.5 & 18.3 & 81.2 & 40.0 & 76.5 & 58.1 & 25.8 & 82.6 & 30.3 & 34.4 & 3.4 & 21.6 & 21.5 & 42.6 \\
		CLAN~\cite{luo2019taking} & 87.0 & 27.1 & 79.6 & 27.3 & 23.3 & 28.3 & 35.5 & 24.2 & 83.6 & 27.4 & 74.2 & 58.6 & 28.0 & 76.2 & 33.1 & 36.7 & 6.7 & 31.9 & 31.4 & 43.2 \\
		APODA~\cite{yang2020adversarial} & 85.6 & 32.8 & 79.0 & 29.5 & 25.5 & 26.8 & 34.6 & 19.9 & 83.7 & \textbf{40.6} & 77.9 & 59.2 & 28.3 & 84.6 & 34.6 & 49.2 & 8.0 & \textbf{32.6} & 39.6 & 45.9 \\
		PatchAlign~\cite{tsai2019domain} & \textbf{92.3} & 51.9 & 82.1 & 29.2 & 25.1 & 24.5 & 33.8 & 33.0 & 82.4 & 32.8 & 82.2 & 58.6 & 27.2 & 84.3 & 33.4 & 46.3 & 2.2 & 29.5 & 32.3 & 46.5 \\
		\hline
		AdvEnt~\cite{vu2019advent} & 89.4 & 33.1 & 81.0 & 26.6 & \textbf{26.8} & 27.2 & 33.5 & 24.7 & 83.9 & 36.7 & 78.8 & 58.7 & 30.5 & 84.8 & \textbf{38.5} & 44.5 & 1.7 & 31.6 & 32.4 & 45.5 \\
		\hline
		Source &  - & - & - & - & - & - & - & - & -& - & - & - & - & - & - & - & - & - & - & 29.2\\
		FCAN~\cite{zhang2018fully} & - & - & - & - & - & - & - & - & -& - & - & - & - & - & - & - & - & - & - & 46.6 \\
		\hline
		Source & 71.3 & 19.2 & 69.1 & 18.4 & 10.0 & 35.7 & 27.3 &  6.8 & 79.6 & 24.8 & 72.1 & 57.6 & 19.5 & 55.5 & 15.5 & 15.1 & 11.7 & 21.1 & 12.0 & 33.8\\
		CBST \cite{zou2018unsupervised} & 91.8 & 53.5 & 80.5 & 32.7 & 21.0 & 34.0 & 28.9 & 20.4 & 83.9 & 34.2 & 80.9 & 53.1 & 24.0 & 82.7 & 30.3 & 35.9 & 16.0 & 25.9 & 42.8 & 45.9\\
		MRKLD \cite{zou2019confidence} & 91.0 & \textbf{55.4} & 80.0 & 33.7 & 21.4 & 37.3 & 32.9 & 24.5 & 85.0 & 34.1 & 80.8 & 57.7 & 24.6 & 84.1 & 27.8 & 30.1 & \textbf{26.9} & 26.0 & 42.3 & 47.1\\
		\hline
		Source & 51.1 & 18.3 & 75.8 & 18.8 & 16.8 & 34.7 & 36.3 & 27.2 & 80.0 & 23.3 & 64.9 & 59.2 & 19.3 & 74.6 & 26.7 & 13.8 & 0.1 & 32.4 & 34.0 & 37.2\\
		MRNet~\cite{zheng2019unsupervised}  & 89.1 & 23.9 & 82.2 & 19.5 & 20.1 & 33.5 & 42.2 & 39.1 & \textbf{85.3} & 33.7 & 76.4 & 60.2 & 33.7 & 86.0 & 36.1 & 43.3 & 5.9 & 22.8 & 30.8 & 45.5 \\
		MRNet+Pseudo & 90.5 & 35.0 & 84.6 & 34.3 & 24.0 & 36.8 & 44.1 & 42.7 & 84.5 & 33.6 & \textbf{82.5} & 63.1 & 34.4 & 85.8 & 32.9 & 38.2 & 2.0 & 27.1 & 41.8 & 48.3 \\
		MRNet+Ours & 90.4 & 31.2 & \textbf{85.1} & \textbf{36.9} & 25.6 & \textbf{37.5} & \textbf{48.8} & \textbf{48.5} & \textbf{85.3} & 34.8 & 81.1 & \textbf{64.4} & \textbf{36.8} & \textbf{86.3} & 34.9 & \textbf{52.2} & 1.7 & 29.0 & \textbf{44.6} & \textbf{50.3} \\
		\shline
	\end{tabular}
	}
	\caption{Quantitative results on GTA5 $\rightarrow$ Cityscapes. We present pre-class IoU and mIoU. The best accuracy in every column is in \textbf{bold}.}
	\label{table:gtacity}
\end{table*}

\begin{table*}[!t]
	\centering
	\resizebox{\linewidth}{!}{
	\begin{tabular}{c|cccccccccccccccc|c|c}
		\shline
		Method & Road & SW & Build & Wall* & Fence* & Pole* & TL & TS & Veg. & Sky & PR & Rider & Car & Bus & Motor & Bike & mIoU* & mIoU\\
		\shline
		\hline
		Source & 55.6 & 23.8 & 74.6 & $-$ & $-$ & $-$ & 6.1 & 12.1 & 74.8 & 79.0 & 55.3 & 19.1 & 39.6 & 23.3 & 13.7 & 25.0 & 38.6 & $-$ \\
		SIBAN~\cite{luo2019significance} & 82.5 & 24.0 & 79.4 & $-$ & $-$ & $-$ & 16.5 & 12.7 & 79.2 & 82.8 & 58.3 & 18.0 & 79.3 & 25.3 & 17.6 & 25.9 & 46.3 & $-$ \\
    	PatchAlign~\cite{tsai2019domain} & 82.4 & 38.0 & 78.6 & 8.7 & 0.6 & 26.0 & 3.9 & 11.1 & 75.5 & 84.6 & 53.5 & 21.6 & 71.4 & 32.6 & 19.3 & 31.7 & 46.5 & 40.0 \\
		AdaptSegNet~\cite{tsai2018learning} & 84.3 & 42.7 & 77.5 & $-$ & $-$ & $-$ & 4.7 & 7.0 & 77.9 & 82.5 & 54.3 & 21.0 & 72.3 & 32.2 & 18.9 & 32.3 & 46.7 & $-$ \\
		CLAN~\cite{luo2019taking} & 81.3 & 37.0 & 80.1 & $-$ & $-$ & $-$ & 16.1 & 13.7 & 78.2 & 81.5 & 53.4 & 21.2 & 73.0 & 32.9 & 22.6 & 30.7 & 47.8 & $-$  \\
		CCM~\cite{li2020content} & 79.6 & 36.4 & 80.6 & 13.3 & 0.3 & 25.5 & 22.4 & 14.9 & 81.8 & 77.4 & 56.8 & 25.9 & 80.7 & 45.3 & \textbf{29.9} &  52.0 & 52.9 & 45.2 \\
		APODA~\cite{yang2020adversarial} & 86.4 & 41.3 & 79.3 & $-$ & $-$ & $-$ & 22.6 & 17.3 & 80.3 & 81.6 & 56.9 & 21.0 & 84.1 & \textbf{49.1} & 24.6 & 45.7 & 53.1 & $-$  \\
		\hline
		AdvEnt \cite{vu2019advent} & 85.6 & 42.2 & 79.7 & 8.7 & 0.4 & 25.9 & 5.4 & 8.1 & 80.4 & \textbf{84.1} & 57.9 & 23.8 & 73.3 & 36.4 & 14.2 & 33.0 & 48.0 & 41.2 \\
		\hline
		Source & 64.3 & 21.3 & 73.1 & 2.4 & 1.1 & 31.4 & 7.0 & 27.7 & 63.1 & 67.6 & 42.2 & 19.9 & 73.1 & 15.3 & 10.5 & 38.9 & 40.3 & 34.9 \\
		CBST \cite{zou2018unsupervised} & 68.0 & 29.9 & 76.3 & 10.8 & 1.4 & 33.9 & 22.8 & 29.5 & 77.6 & 78.3 & 60.6 & 28.3 & 81.6 & 23.5 & 18.8 & 39.8 & 48.9 & 42.6 \\
		MRKLD \cite{zou2019confidence} & 67.7 & 32.2 & 73.9 & 10.7 & 1.6 & \textbf{37.4} & 22.2 & \textbf{31.2} & 80.8 & 80.5 & 60.8 & \textbf{29.1} & 82.8 & 25.0 & 19.4 & 45.3 & 50.1 & 43.8 \\
		\hline
		Source & 44.0 & 19.3 & 70.9 & 8.7 & 0.8 & 28.2 & 16.1 & 16.7 & 79.8 & 81.4 & 57.8 & 19.2 & 46.9 & 17.2 & 12.0 & 43.8 & 40.4 & 35.2 \\
		MRNet~\cite{zheng2019unsupervised} & 82.0 & 36.5 & 80.4 & 4.2 & 0.4 & 33.7 & 18.0 & 13.4 & 81.1 & 80.8 & 61.3 & 21.7 & 84.4 & 32.4 & 14.8 & 45.7 & 50.2 & 43.2 \\
		MRNet+Pseudo & 83.1 & 38.2 & 81.7 & 9.3 & 1.0 & 35.1 & 30.3 & 19.9 & \textbf{82.0} & 80.1 & 62.8 & 21.1 & 84.4 & 37.8 & 24.5 & \textbf{53.3} & 53.8 & 46.5 \\
		MRNet+Ours & \textbf{87.6} & 41.9 & \textbf{83.1} & \textbf{14.7} & \textbf{1.7} & 36.2 & \textbf{31.3} & 19.9 & 81.6 & 80.6 & \textbf{63.0} & 21.8 & \textbf{86.2} & 40.7 & 23.6 & 53.1 & \textbf{54.9} & \textbf{47.9} 
		\\
		\shline
	\end{tabular}
	}
	\caption{Quantitative results on SYNTHIA $\rightarrow$ Cityscapes. We present pre-class IoU, mIoU and mIoU*. mIoU and mIoU* are averaged over 16 and 13 categories, respectively. The best accuracy in every column is in \textbf{bold}.}
	\label{table:syncity}
\end{table*}

\begin{table} [!t]
	\centering
    \resizebox{\linewidth}{!}{
	\begin{tabular}{l|ccccccccc|c}
		\shline
		Method & \rotatebox{90}{road} & \rotatebox{90}{sidewalk} & \rotatebox{90}{building} & \rotatebox{90}{light} & \rotatebox{90}{sign} & \rotatebox{90}{sky} & \rotatebox{90}{person} & \rotatebox{90}{automobile} & \rotatebox{90}{two-wheel} & mIoU\\
		
		\shline
		
		Source & 79.2 & 49.3 & 73.1 & 55.6 & 37.3 & 36.1 & 54.0 & 81.3 & 49.7 & 61.9 \\
		AdaptSegNet~\cite{tsai2018learning}  & 95.1 & 64.0 & 75.7 & 61.3 & 35.5 & 63.9 & 58.1 & 84.6 & 57.0 & 69.5 \\
		PatchAlign~\cite{tsai2019domain} & 94.4 & 63.5 & 82.0 & 61.3 & 36.0 & 76.4 & 61.0 & 86.5 & 58.6 & 72.0 \\
		\hline
		MRNet~\cite{zheng2019unsupervised} & \textbf{95.9} & 73.5 & 86.2 & 69.3 & 31.9 & 87.3 & 57.9 & 88.8 & \textbf{61.5} & 72.5 \\
		MRNet+Pseudo & 95.1 & 72.5 & 87.0 & 72.2 & 37.4 & 87.9 & \textbf{63.4} & \textbf{90.5} & 58.9 & 73.9 \\
		MRNet+Ours & \textbf{95.9} & \textbf{73.7} & \textbf{87.4} & \textbf{72.8} & \textbf{43.1} & \textbf{88.6} & 61.7 & 89.6 & 57.0 & \textbf{74.4} \\
		\shline
	\end{tabular}}
	\caption{
		Quantitative results on the cross-city benchmark: Cityscapes $\rightarrow$ Oxford RobotCar. The best accuracy in every column is in \textbf{bold}.
	}
		\label{table:oxford}
\end{table}

\subsection{Comparisons with state-of-the-art methods}
\textbf{Synthetic-to-real.} We compare the proposed method with other recent semantic segmentation adaptation methods that have reported the results or can be re-implemented by us on three benchmarks. For a fair comparison, we mainly compare the results based on the same network structure, \ie, DeepLabv2. The competitive methods cover a wide range of approaches and could be roughly categorised according to the usage of pseudo label: 
AdaptSegNet \cite{tsai2018learning}, SIBAN \cite{luo2019significance}, CLAN \cite{luo2019taking}, APODA \cite{yang2020adversarial} and PatchAlign \cite{tsai2018learning} do not leverage the pseudo labels and focus on aligning the distribution between the source domain and the target domain; CBST \cite{zou2018unsupervised}, MRKLD \cite{zou2019confidence},  and our implemented MRNet+Pseudo are based on the pseudo label learning to fully exploit the unlabeled target-domain data.

 First of all, we consider the widely-used GTA5 $\rightarrow$ Cityscapes benchmark.  Table \ref{table:gtacity} shows that: (1) The proposed method arrives the state-of-the-art results $50.3\%$ mIoU, which surpasses other methods. Besides, the proposed method also yields the competitive performance in terms of the pre-class IoU. 
(2) Comparing to our baseline, \ie, MRNet+Pseudo ($48.3\%$ mIoU), which adopts the conventional pseudo learning, the proposed method ($50.3\%$ mIoU) gains $+2.0\%$ mIoU improvement. It verifies the effectiveness of the proposed method in rectifying the learning from the noisy pseudo label. The variance regularization plays an important role in achieving this result; 
 (3) Meanwhile, we could observe that the proposed method outperforms the source-domain model, \ie, MRNet ($45.5\%$ mIoU), which provides the pseudo label, $4.8$ mIoU. It verifies the effectiveness of the pseudo label learning that push the model to be confident about the prediction. If most pseudo labels are correct, the pseudo label learning could effectively boost the target-domain performance. 
 (4) The proposed method also surpasses the other domain alignment method by a relatively large margin. For example, the modified AdaptSegNet, \ie, PatchAlign \cite{tsai2018learning}, leverages the patch-level information, yielding $46.5\%$, which is inferior to ours.
 (5) Without using the prior knowledge, the proposed method is also superior to other pseudo label learning works, \ie, CBST \cite{zou2018unsupervised} and MRKLD \cite{zou2019confidence}. CBST \cite{zou2018unsupervised} introduces the location knowledge, \eg, sky is always in the upper bound of the image. In this work, we do not apply such prior knowledge, but we note that the prior knowledge is compatible with our method.

We observe a similar result on SYNTHIA $\rightarrow$ Cityscapes (see  Table~\ref{table:syncity}). Following the setting in \cite{zou2018unsupervised,zou2019confidence}, we include the mIoU results of 13 categories as well as 16 categories, which also calculate IoU of other three small-scale objectives, \ie, Wall, Fence and Pole. The proposed method has achieved $47.9$ mIoU of 16 categories and $54.9$ mIoU$^*$ of 13 categories. Comparing to the baseline, MRNet+Pseudo, we yield $+1.4\%$ mIoU and $+1.1\%$ mIoU$^{*}$ improvement. Meanwhile, the proposed method also outperforms the second best method, \ie, APODA \cite{yang2020adversarial}, $1.8\%$ mIoU$^{*}$. 

\noindent\textbf{Cross-city.} We further evaluate the proposed method on the cross-city benchmark, \ie, Cityscapes $\rightarrow$ Oxford RobotCar. Both of the source-domain and target-domain datasets are collected in the real-world scenario. We follow the settings in \cite{tsai2019domain} to report IoU of the shared 9 categories between the two datasets. As shown in Table \ref{table:oxford}, the proposed method arrives $74.4\%$ mIoU. Comparing to the baseline, \ie, MRNet+Pseudo ($73.9\%$), the improvement ($+0.5\%$) on the cross-city benchmark is relatively limited. 
Therefore, the baseline, MRNet+Pseudo, also could obtain competitive results by directly utilizing all pseudo labels. Besides, it is worthy to note that the proposed method has arrived the 6 of 9 best pre-class IoU accuracy, and achieved $+5.7\%$ on the class of traffic sign, which is a small-scale objective. 

\begin{figure*}[t]
\begin{center}
     \includegraphics[width=1\linewidth]{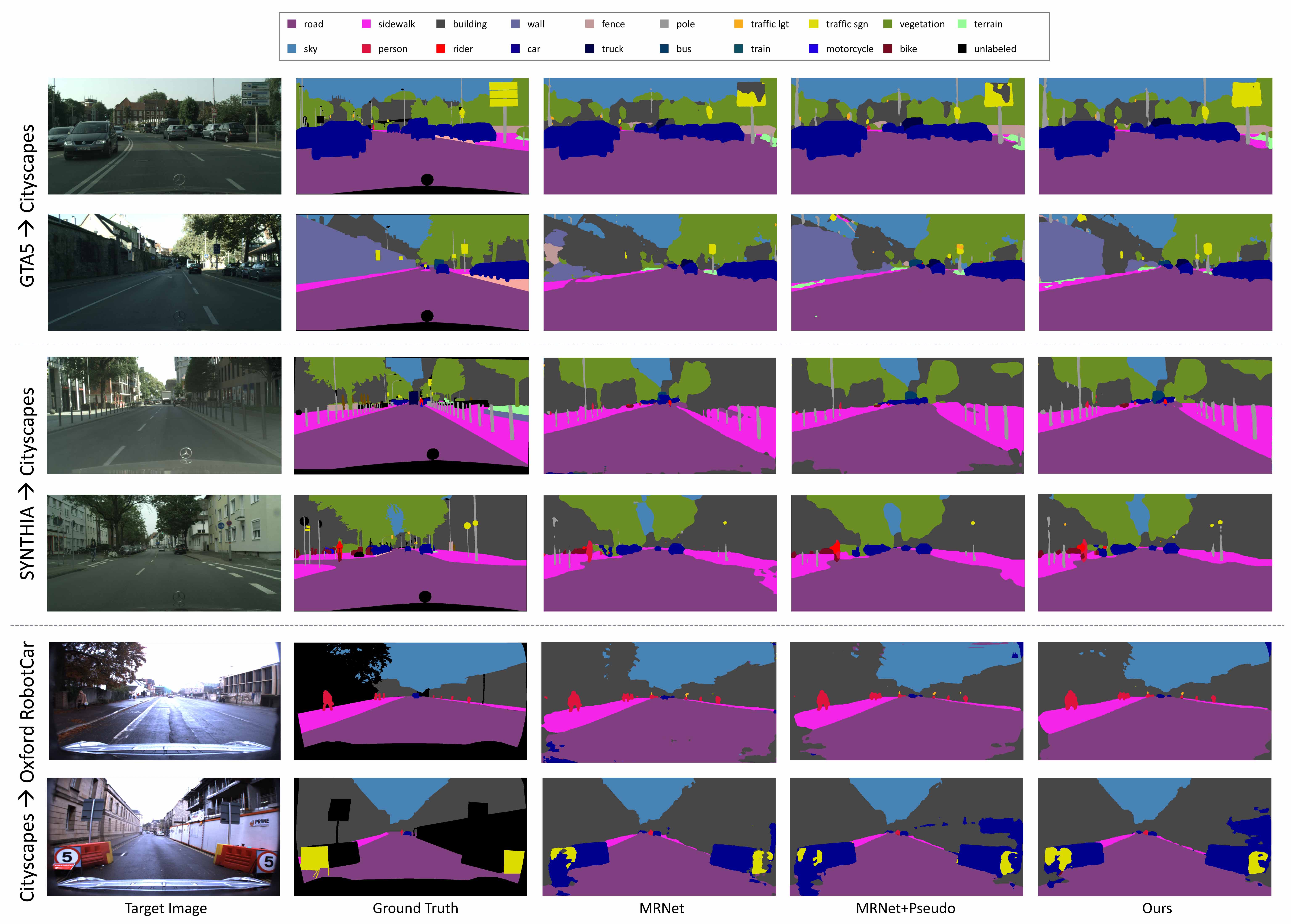}
\end{center} 
      \caption{  Qualitative results of semantic segmentation adaptation on GTA5 $\rightarrow$ Cityscapes, SYNTHIA $\rightarrow$ Cityscapes and Cityscapes $\rightarrow$ Oxford RobotCar. We show the original target image, the ground-truth segmentation, the output of the source model, \ie, MRNet, and the baseline, \ie, MRNet+Pseudo. Our results are in the right column. (Best viewed in \emph{color}). }
      \label{fig:result}
\end{figure*}

\noindent\textbf{Visualization.} As shown in Figure~\ref{fig:result}, we provide the qualitative results of the semantic segmentation adaptation on all three benchmarks. Comparing to the source model, the pseudo label learning could significantly improve the performance. Besides, in contrast with the baseline method with conventional pseudo label learning, we observe that the proposed variance regularization has better scalability to small-scale objectives, such as traffic signs and poles. 
It is because that the noisy pseudo label usually contains the error of predicting the rare category to the common category, \ie, large-scale objectives. The proposed method rectifies the learning from such mistakes, yielding more reasonable segmentation prediction. 

\subsection{Further Evaluations}
\textbf{Variance Regularization vs. Handcrafted Threshold.} The proposed variance regularization is free from setting the threshold. To verify the effectiveness of the variance regularization, we also compare the conventional pseudo label learning with different thresholds. As shown in Table \ref{table:threshold}, the proposed regularization arrives the superior performance to the hand-crafted threshold. 
It is due to that the variance regularization could be viewed as a dynamic threshold, providing different thresholds for different pixels in the same image. For the coherent predictions, the model is prone to learning the pseudo label and maximizing the impact of such labels. 
For the incoherent results, the model is prone to neglecting the pseudo label automatically and minimizing the negative effect of noisy labels. 
The best handcrafted threshold is to neglect the label with the prediction score $\leq0.90$, yielding $48.4\%$ mIoU. In contrast, the proposed method achieves $50.3\%$ mIoU with $+1.9\%$ increment. 

\setlength{\tabcolsep}{10pt}
\begin{table}
\footnotesize
\begin{center}
\begin{tabular}{l|c|c}
\shline
Methods  & Threshold & mIoU \\
\shline
MRNet \cite{zheng2019unsupervised}  & - &  45.5 \\
\hline
Pseudo Learning & $>0.99$ & 45.5 \\
Pseudo Learning & $>0.95$ & 47.2 \\
Pseudo Learning & $>0.90$ & 48.4 \\
Pseudo Learning & $>0.80$ & 48.1 \\
Pseudo Learning & $>0.70$ & 48.2 \\
Pseudo Learning & $>0.00$ & 48.3 \\
\hline
Ours & - & 50.3 \\
\shline
\end{tabular}
\end{center}
\caption{Variance Regularization vs. Handcrafted Threshold. The proposed method is free from hand-crafted threshold. `$>k$' denotes that we only utilize the label confidence $>k$ to train the model. We report the mIoU accuracy on GTA5 $\rightarrow$ Cityscapes.}
\label{table:threshold}
\end{table}

\begin{figure*}[t]
\begin{center}
     \includegraphics[width=1\linewidth]{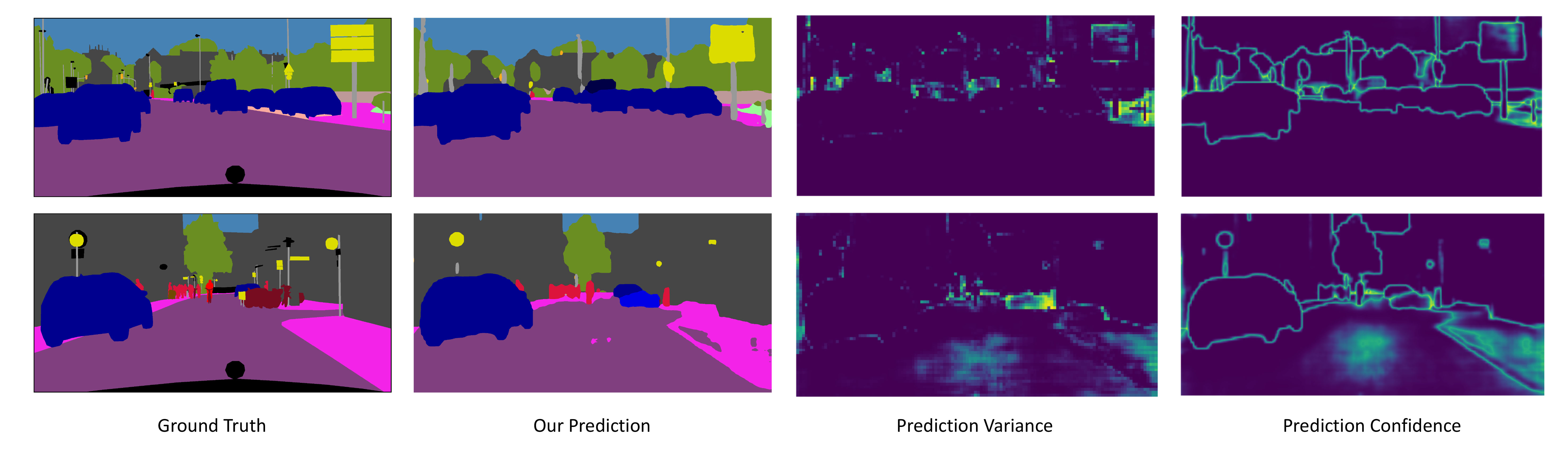}
\end{center} 
      \caption{  Qualitative results of the discrepancy between the prediction variance and the prediction confidence. We could observe that the prediction variance used in this work has more overlaps with the ambiguous predictions, while the prediction confidence usually focuses on the edge of the two different classes. (Best viewed in \emph{color}). }
      \label{fig:uncertain}
\end{figure*}

\begin{figure}[t]
\begin{center}
     \includegraphics[width=1\linewidth]{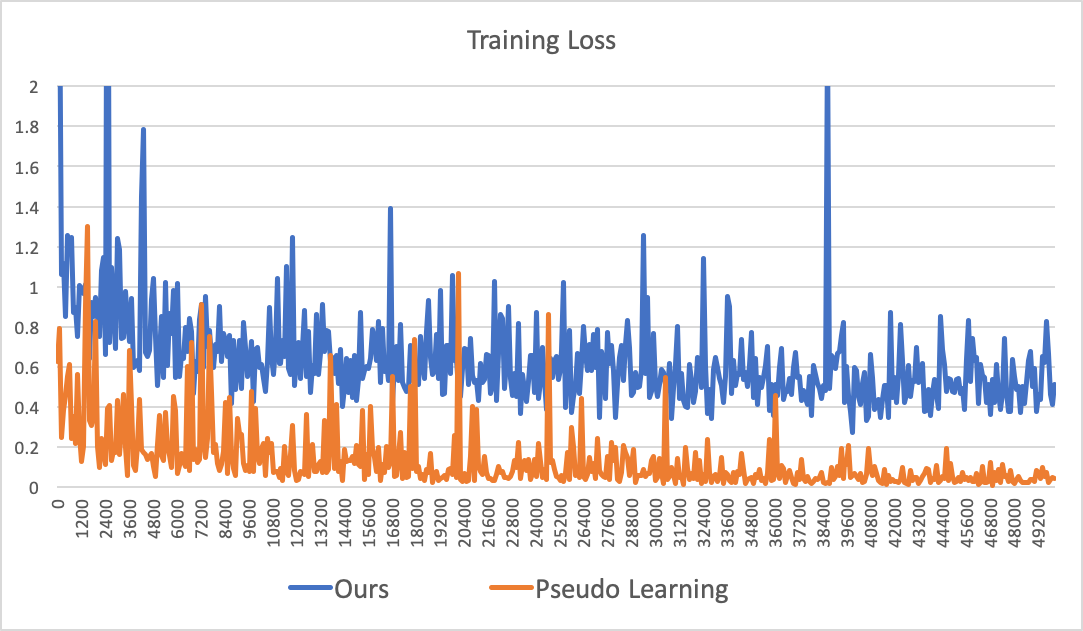}
\end{center} 
      \caption{ The training loss of the proposed method and the pseudo label learning. The pseudo label learning is prone to over-fit all pseudo label, and the training loss converges to zero. In contrast, the proposed method would converge to one non-zero constant while training. }
      \label{fig:loss}
\end{figure}

\noindent\textbf{Could the proposed method work on the pseudo label generated by other models (\eg, with more noise)?} To verify the scalability of the proposed method, we adopt the AdaptSegNet \cite{tsai2018learning} to generate pseudo labels. AdaptSegNet is inferior to MRNet in terms of the mIoU on GTA5 $\rightarrow$ Cityscapes. As shown in Table~\ref{table:poorlabel}, the proposed method still could learn from the label generated by AdaptSegNet, improving the performance from $42.4\%$ to $47.4\%$. Meanwhile, the proposed method is also superior to the baseline method with the conventional pseudo learning ($46.8\%$ mIoU).

\setlength{\tabcolsep}{10pt}
\begin{table}
\footnotesize
\begin{center}
\begin{tabular}{l|c|c}
\shline
Methods  & Pseudo Label & mIoU \\
\shline
AdaptSegNet~\cite{tsai2018learning}  & - &  42.4 \\
\hline
Pseudo Learning & AdaptSegNet & 46.8 \\
Ours   & AdaptSegNet &  47.4 \\
\shline
MRNet~\cite{zheng2019unsupervised}  & - &  45.5 \\
\hline
Pseudo Learning & MRNet & 48.3 \\
Ours   & MRNet & 50.3 \\
\shline
\end{tabular}
\end{center}
\caption{Ablation study of the impact of different pseudo labels. The model name in the `Pseudo Label` column denotes that we deploy the pseudo label generated by the corresponding model.}
\label{table:poorlabel}
\end{table}

\setlength{\tabcolsep}{20pt}
\begin{table}
\small
\begin{center}
\begin{tabular}{l|c}
\shline
Dropout Rate & mIoU \\
\shline
Pseudo Learning & 48.3 \\
\hline
droprate = 0 & 49.6 \\
droprate = 0.1  & 50.3 \\
droprate = 0.3  & 50.1 \\
droprate = 0.5  & 50.1 \\
droprate = 0.7  & 50.0 \\
\shline
\end{tabular}
\end{center}
\caption{Ablation study of dropout rate on GTA5 $\rightarrow$ Cityscapes.}
\label{table:dropout}
\end{table}

\setlength{\tabcolsep}{7pt}
\begin{table}
\scriptsize
\begin{center}
\begin{tabular}{l|c|c|c}
\shline
Methods & Right-prediction & Wrong-prediction & Uncertainty \\
& Certainty & Certainty & Gap\\
\shline
MC-dropout 0.5 & 0.9945 & 0.9733 & 0.0212 \\
MC-dropout 0.7 & 0.9870 & 0.9396 & 0.0474 \\
MC-dropout 0.9 & 0.9486 & 0.8118 & 0.1368 \\
\hline
Ours & 0.9767 & 0.8410 & 0.1357 \\
Ours + dropout 0.5 & 0.9673 & 0.8065 & \textbf{0.1608} \\
\shline
\end{tabular}
\end{center}
\caption{Comparison with Monte Carlo Dropout.}
\label{table:mc-drop}
\end{table}

\setlength{\tabcolsep}{15pt}
\begin{table}
\footnotesize
\begin{center}
\begin{tabular}{l|c}
\shline
Distance Functions & mIoU \\
\shline
$\E [(F(x_t^j|\theta_t) - F_{aux}(x_t^j|\theta_t) )^2] $ & 49.6 \\
$\E [F_{aux}(x_t^j|\theta_t) \log (\frac{F_{aux}(x_t^j|\theta_t)}{F(x_t^j|\theta_t)})]$ & 49.4 \\
\hline
$\E [F(x_t^j|\theta_t) \log (\frac{F(x_t^j|\theta_t)}{F_{aux}(x_t^j|\theta_t)})]$ & 50.3 \\
\shline
\end{tabular}
\end{center}
\caption{Ablation study of distance functions on GTA5 $\rightarrow$ Cityscapes.}
\label{table:function}
\end{table}

\setlength{\tabcolsep}{15pt}
\begin{table}
\small
\begin{center}
\begin{tabular}{l|c|c}
\shline
$\alpha$ & $\beta$ & mIoU\\
\shline
1.0 & 0.0 & 49.3 \\
0.0 & 1.0 & 47.8 \\
1.0 & 1.0 & 50.1 \\
\hline
1.0 & 0.5 & 50.3 \\
\shline
\end{tabular}
\end{center}
\caption{Sensitivity of inference weighting.}
\label{table:inference}
\end{table}

\noindent\textbf{Training Convergence.} 
As shown in Figure~\ref{fig:loss}, the conventional pseudo label learning (orange line) is prone to over-fit all pseudo labels, including the noisy label. Therefore, the training loss converges to zero. In contrast, the proposed method (blue line) also converges, but does not force the loss to be zero. It is because that we provide the variance regularization term, which could punish the wrong prediction for the uncertain pseudo labels with flexibility. 

\noindent\textbf{Effect of Dropout.} 
The proposed method is not very sensitive to the dropout rate. As shown in Table ~\ref{table:dropout}, we could observe two points: 1) The dropout function is not the main reason for variance of the predictions. Without dropout function ($p=0$), the proposed method still could achieve $49.6\%$ mIoU, which is better than the conventional pseudo label learning. 2) With a propose dropout rate, the proposed method could generally achieve  better results around $50\%$ mIoU.

\noindent\textbf{Uncertainty of High-confidence Predictions.} 
We analyze the variance of high-confidence predictions on Cityscape. Specifically, we calculate the average uncertainty of right-assigned and wrong-assigned prediction with a confidence score$> 0.95$. Here we use the metric $exp\{-D_{kl}\}$ in Equation \ref{eq:rectified} to report the variance value. The high value means low uncertainty. 
The average variance of right-assigned high-confidence labels is $0.9901$, when the average variance of wrong-assigned high-confidence labels is $0.9332$. We could see one significant variance gap between the right-assigned labels and wrong-assigned labels, even if they all achieve a high confidence score. The result verifies that the variance value could reflect the difference between wrong-assigned labels and right-assigned labels.

\noindent\textbf{Comparison with Monte Carlo Dropout.} 
Monte Carlo Dropout (MC-Dropout)~\cite{gal2016dropout} activates the dropout function when inference to obtain various predictions. Here we compare the ability of representing the uncertainty of the proposed method and MC-Dropout. 
For a fair comparison, we just replace the prediction of the aux classifier with the main classifier $F_{drop}$ with MC dropout rate of $\{0.5, 0.7, 0.9\}$.
\begin{equation}
D_{mc} = \E [F(x_t^j|\theta_t) \log (\frac{F(x_t^j|\theta_t)}{F_{drop}(x_t^j|\theta_t)})].
\end{equation}
Since the prediction score could not reflect the ground-truth uncertainty, we introduce one new metric called uncertainty gap as indicator. Uncertainty gap is the variance difference of right predictions and wrong predictions. Generally, we hope that the right prediction obtains low uncertainty value, while the wrong prediction obtains high uncertainty value. In practice, we use the $exp(-D)$ to keep the value in [0,1]. 
As shown in Table \ref{table:mc-drop}, the proposed method obtains 0.1357 variance gap, which is competitive to MC-dropout with 0.9 drop rate. The proposed method is also complementary to MC-dropout. The proposed method with MC-dropout could further boost the uncertainty gap.  Meanwhile, it is worth noting that the proposed method directly leverages the variance of both main and auxiliary classifiers without multiple inferences, which can largely save the test time.

\noindent\textbf{Effect of Distance Functions.} In fact, KL-divergence is an alternative option for variance calculation. We could swap the main and aux classifiers to calculate the distance or use mean-square error (MSE). Here we add one experiment to compare common distance functions (see Table~\ref{table:function}). First, we could observe that the model is not very sensitive to the distance metric, since the performances are close. Second, the KL-divergence used in Method is slightly better than swapping the predictions and MSE distance.

\noindent\textbf{Effect of Inference Weighting.} 
Inference weighting is one practical trick to combine the predictions of both main and auxiliary classifiers. Generally, the main classifier could achieve better performance, so we give the prediction of the main classifier a larger weight of $\alpha = 1$ and assign $\beta = 0.5$ to the prediction of auxiliary classifier. $Output = \argmax( \alpha F(x_t^j|\theta_t) + \beta F_{aux}(x_t^j|\theta_t))$. This trick could slightly improve the final performance. Here we provide the ablation study on the sensitivity of inference weighting in Table \ref{table:inference}. If we only deploy the main classifier ($ \alpha=1, \beta= 0$), the model could achieve $49.3\%$ mIoU accuracy. When we combine the prediction of two classifiers, the performance could be improved about $1.0\%$ mIoU.

\noindent\textbf{Uncertainty Visualization.} As a by-product, we also could estimate the prediction uncertainty when inference. We provide the visualization results to show the difference between the uncertainty estimation and the confidence score. As shown in Figure~\ref{fig:uncertain}, we observe that the model is prone to provide the low confidence score of the boundary pixels, which does not provide the effective cue to the ambiguous prediction. Instead, the proposed prediction variance reflects the label uncertainty, and the highlight area in prediction variance map has lots of overlaps with the wrong prediction.  

\section{Conclusion} \label{conclusion}
We identify the challenge of pseudo label learning in adaptive semantic segmentation and present a simple and effective method to estimate the prediction uncertainty during training. We also involve the uncertainty into the optimization objective as the variance regularization to rectify the training. The regularization helps the model learn from the noisy label, without introducing extra parameters or modules. As a result, we achieve the competitive performance on three benchmarks, including two synthetic-to-real benchmarks and one cross-city benchmark. In the future, we will continue to investigate the usage of uncertainty and the applications to other related tasks, \eg, medical imaging.

{\footnotesize
\bibliographystyle{apalike}
\bibliography{egbib}
}

\end{document}